\def\year{2018}\relax
\documentclass[letterpaper]{article} 
\usepackage{aaai18}  
\usepackage{times}  
\usepackage{helvet}  
\usepackage{courier}  
\usepackage{url}  
\usepackage{graphicx}  
\usepackage{multirow}
\usepackage{color}
\usepackage{amsthm}
\usepackage{amssymb}
\usepackage{amsmath}
\usepackage{arydshln}
\usepackage{amsmath}
\usepackage{algorithm}
\usepackage[noend]{algpseudocode}
\usepackage{algorithmicx}
\usepackage{microtype}
\usepackage{stfloats}
\usepackage{makecell}
\usepackage{MnSymbol}
\usepackage[shortlabels]{enumitem}
\usepackage[makeroom]{cancel}

\newcommand{\pipeline}{Multi-attention Recurrent Network}
\newcommand{\pipelines}{MARN}

\newcommand{\mab}{Multi-attention Block}
\newcommand{\mabs}{MAB}
\newcommand{\intra}{view-specific}
\newcommand{\inter}{cross-view}

\newcommand{\ourl}{Long-short Term Hybrid Memory}
\newcommand{\ours}{LSTHM}
\newcommand{\tname}{Long-short Term Hybrid Memory}
\newcommand{\tnames}{LSTHM}

\usepackage{booktabs}

\newcommand{\commentAP}[1]{\textcolor{blue}{Amir to Paul: #1}}

\makeatletter
\newcommand\footnoteref[1]{\protected@xdef\@thefnmark{\ref{#1}}\@footnotemark}
\makeatother

\begin{document}

%

\title{\pipeline \ for Human Communication Comprehension}

\author{
Amir Zadeh\\
Carnegie Mellon University, USA \\
{\tt abagherz@cs.cmu.edu} \\ \And
Paul Pu Liang\\
Carnegie Mellon University, USA \\
{\tt pliang@cs.cmu.edu} \\ \And
Soujanya Poria \\
NTU, Singapore \\
{\tt sporia@ntu.edu.sg} \\ \AND
Prateek Vij \\
NTU, Singapore \\
{\tt prateek@sentic.net} \\ \And
Erik Cambria \\
NTU, Singapore \\
{\tt cambria@ntu.edu.sg} \\ \And
Louis-Philippe Morency \\
Carnegie Mellon University, USA\\
{\tt morency@cs.cmu.edu}\\
}
\maketitle

\begin{abstract}
Human face-to-face communication is a complex multimodal signal. We use words (language modality), gestures (vision modality) and changes in tone (acoustic modality) to convey our intentions. Humans easily process and understand face-to-face communication, however, comprehending this form of communication remains a significant challenge for Artificial Intelligence (AI). AI must understand each modality and the interactions between them that shape human communication. In this paper, we present a novel neural architecture for understanding human communication called the \pipeline \ (\pipelines). The main strength of our model comes from discovering interactions between modalities through time using a neural component called the \mab \ (\mabs) and storing them in the hybrid memory of a recurrent component called the \tname \ (\tnames). We perform extensive comparisons on six publicly available datasets for multimodal sentiment analysis, speaker trait recognition and emotion recognition. \pipelines \ shows state-of-the-art performance on all the datasets. 
\end{abstract}

\section{Introduction}
Humans communicate using a highly complex structure of multimodal signals. We employ three modalities in a coordinated manner to convey our intentions: \textit{language modality} (words, phrases and sentences), \textit{vision modality} (gestures and expressions), and \textit{acoustic modality} (paralinguistics and changes in vocal tones) \cite{morency2011towards}. Understanding this multimodal communication is natural for humans; we do it subconsciously in the cerebrum of our brains everyday. However, giving Artificial Intelligence (AI) the capability to understand this form of communication the same way humans do, by incorporating all involved modalities, is a fundamental research challenge. Giving AI the capability to understand human communication narrows the gap in computers' understanding of humans and opens new horizons for the creation of many intelligent entities.

The coordination between the different modalities in human communication introduces \intra \ and \inter \ dynamics \cite{tensoremnlp17}. View-specific dynamics refer to dynamics within each modality independent of other modalities. For example, the arrangement of words in a sentence according to the generative grammar of the language (language modality) or the activation of facial muscles for the presentation of a smile (vision modality). Cross-view dynamics refer to dynamics between modalities and are divided into synchronous and asynchronous categories. An example of synchronous \inter \ dynamics is the simultaneous co-occurrence of a smile with a positive sentence and an example of asynchronous \inter \ dynamics is the delayed occurrence of a laughter after the end of sentence. For machines to understand human communication, they must be able to understand these \intra \ and \inter \ dynamics. 

To model these dual dynamics in human communication, we propose a novel deep recurrent neural model called the \pipeline \ (\pipelines). \pipelines \ is distinguishable from previous approaches in that it explicitly accounts for both \intra \ and \inter \ dynamics in the network architecture and continuously models both dynamics through time. In \pipelines, \intra \ dynamics within each modality are modeled using a \tname \ (\tnames) assigned to that modality. The hybrid memory allows each modality's \tnames \ to store important \inter \ dynamics related to that modality. Cross-view dynamics are discovered at each recurrence time-step using a specific neural component called the \mab \ (\mabs). The \mabs \ is capable of simultaneously finding multiple \inter \ dynamics in each recurrence timestep. The \pipelines \ resembles the mechanism of our brains for understanding communication, where different regions independently process and understand different modalities \cite{KUZMANOVIC2012179,Sergent55} -- our \tnames \ -- and are connected together using neural links for multimodal information integration \cite{Jiang16064} -- our \mabs. We benchmark \pipelines \ by evaluating its understanding of different aspects of human communication covering sentiment of speech, emotions conveyed by the speaker and displayed speaker traits. We perform extensive experiments on 16 different attributes related to human communication on public multimodal datasets. Our approach shows state-of-the-art performance in modeling human communication for all datasets. 

\section{Related Work}
\label{Related Work}
Modeling multimodal human communication has been studied previously. Past approaches can be categorized as follows:

\textit{Non-temporal Models}: Studies have focused on simplifying the temporal aspect of  \inter \ dynamics \cite{contextmultimodalacl2017,perez2013utterance,wollmer2013youtube} in order to model co-occurrences of information across the modalities. In these models, each modality is summarized in a representation by collapsing the time dimension, such as averaging the modality information through time \cite{abburi2016multimodal}. While these models are successful in understanding co-occurrences, the lack of temporal modeling is a major flaw as these models cannot deal with multiple contradictory evidences, eg. if a smile and frown happen together in an utterance. Furthermore, these approaches cannot accurately model long sequences since the representation over long periods of time become less informative. 

\textit{Early Fusion}: Approaches have used multimodal input feature concatenation instead of modeling \intra \ and \inter \ dynamics explicitly. In other words, these approaches rely on generic models (such as Support Vector Machines or deep neural networks) to learn both \intra \ and \inter \ dynamics without any specific model design. This concatenation technique is known as early fusion \cite{wang2016select,poria2016convolutional}. Often, these early fusion approaches remove the time factor as well \cite{zadeh2016multimodal,morency2011towards}. We additionally compare to a stronger recurrent baseline that uses early fusion while maintaining the factor of time. A shortcoming of these models is the lack of detailed modeling for \intra \ dynamics, which in turn affects the modeling of \inter \ dynamics, as well as causing overfitting on input data \cite{xu2013survey}.

\textit{Late Fusion}: Late fusion methods learn different models for each modality and combine the outputs using decision voting \cite{torstenptsd,Nojavanasghari:2016:DMF:2993148.2993176}. While these methods are generally strong in modeling \intra \ dynamics, they have shortcomings for \inter \ dynamics since these inter-modality dynamics are normally more complex than a decision vote. As an example of this shortcoming, if a model is trained for sentiment analysis using the vision modality and predicts negative sentiment, late fusion models have no access to whether this negative sentiment was due to a frowning face or a disgusted face.

\textit{Multi-view Learning}: Extensions of Hidden Markov Models \cite{baum1966statistical} and Hidden Conditional Random Fields \cite{Quattoni:2007:HCR:1313053.1313265,morency2007latent} have been proposed for learning from multiple different views (modalities) \cite{song2012multi,song2013action}. Extensions of LSTMs have also been proposed in a multi-view setting \cite{rajagopalan2016extending}.

\pipelines \ is different from the first category since we model both \intra \ and \inter \ dynamics. It is differs from the second and third category since we explicitly model \intra \ dynamics using a \tnames \ for each modality as well as \inter \ dynamics using the \mabs. Finally, \pipelines \ is different from the fourth category since it explicitly models \intra \ dynamics and proposes more advanced temporal modeling of \inter \ dynamics.

\section{\pipelines \ Model}
\begin{figure}[t!]
\centering{
\includegraphics[width=\linewidth]{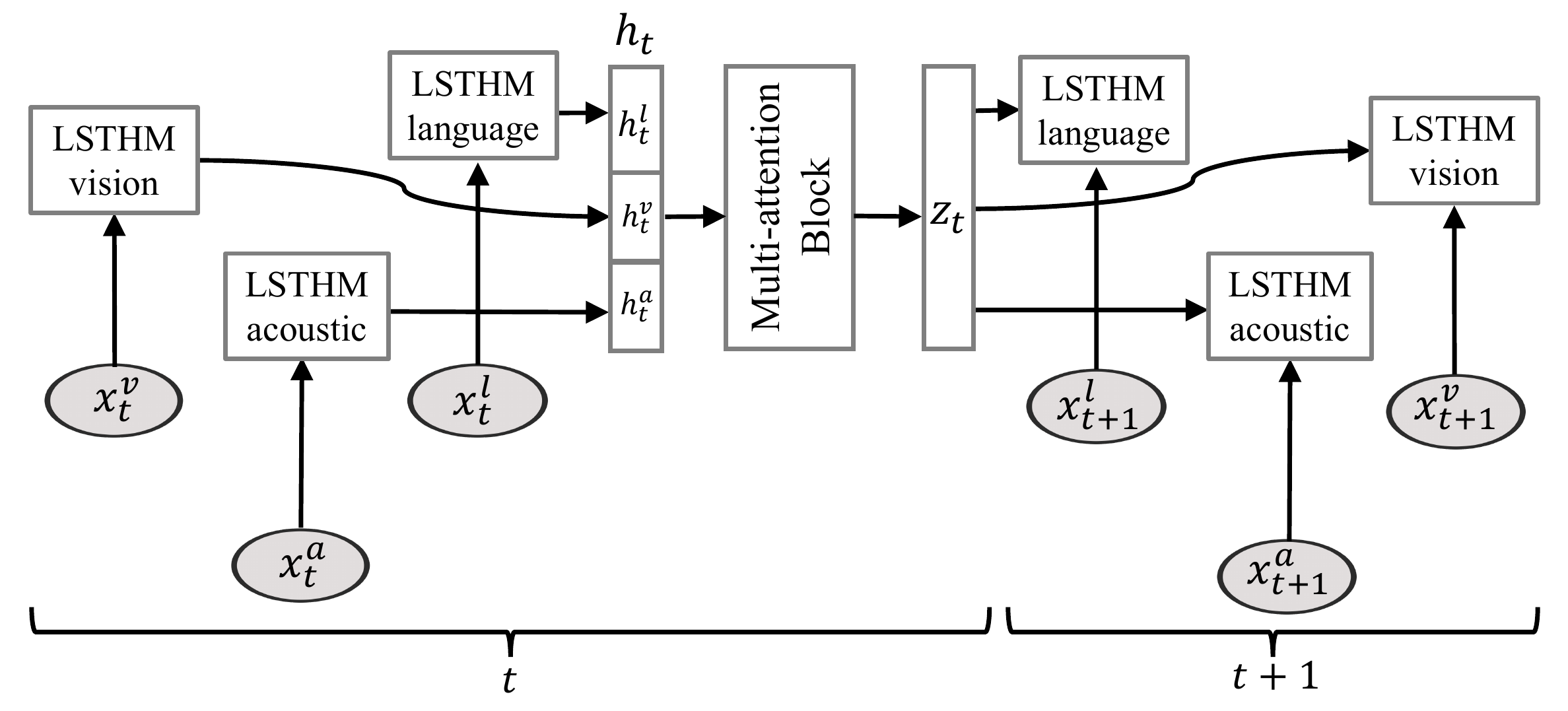}}
    \caption{Overview of \pipeline \ (\pipelines) with \tname \ (\tnames) and \mab \ (\mabs) components, for $M = \{l,v,a\}$ representing the language, vision and acoustic modalities respectively.}
    \label{fig:gmodel}
\end{figure}

In this section we outline our pipeline for human communication comprehension: the \pipeline \ (\pipelines). \pipelines \ has two key components: \tname \ and \mab.  \tname \ (\tnames) is an extension of the Long-short Term Memory (LSTM) by reformulating the memory component to carry hybrid information. \tnames \ is intrinsically designed for multimodal setups and each modality is assigned a unique \tnames. \tnames \ has a hybrid memory that stores \intra \ dynamics of its assigned modality and \inter \ dynamics related to its assigned modality. The component that discovers \inter \ dynamics across different modalities is called the \mab \ (\mabs). The \mabs \ first uses information from hidden states of all \tnames s at a timestep to regress coefficients to outline the multiple existing \inter \ dynamics among them. It then weights the output dimensions based on these coefficients and learns a neural \inter \ dynamics code for \tnames s to update their hybrid memories. Figure \ref{fig:gmodel} shows the overview of the \pipelines. \pipelines \ is differentiable end-to-end which allows the model to be learned efficiently using gradient decent approaches. In the next subsection, we first outline the \tname. We then proceed to outline the \mab \ and describe how the two components are integrated in the \pipelines.

\subsection{\tname}
Long-short Term Memory (LSTM) networks have been among the most successful models in learning from sequential data \cite{hochreiter1997long}. The most important component of the LSTM is a memory which stores a representation of its input through time. In the \tnames \ model, we seek to build a memory mechanism for each modality which in addition to storing \intra \ dynamics, is also able to store the \inter \ dynamics that are important for that modality. This allows the memory to function in a hybrid manner. 

The \tname \ is formulated in Algorithm 1. Given a set of $M$ modalities in the domain of the data, subsequently $M$ \tnames s are built in the \pipelines \ pipeline. For each modality $m \in M$, the input to the $m$th \tnames \ is of the form $\mathbf{X}^m=\{{x}_{1}^m, {x}_{2}^m, {x}_{3}^m, \cdots, {x}_{T}^m \ ; {x}_{t}^m \in \mathbb{R}^{d_{in}^m} \}$, where ${x}^m_{t}$ is the input at time $t$ and $d^m_{in}$ is the dimensionality of the input of modality $m$. For example if $m=l \textrm{(language)}$, we can use word vectors with $d^{l}_{in}=300$ at each time step $t$. $d^{m}_{mem}$ is the dimensionality of the memory for modality $m$. $\sigma$ is the (hard-)sigmoid activation function and $tanh$ is the tangent hyperbolic activation function. $\bigoplus$ denotes vector concatenation and $\odot$ denotes element-wise multiplication. Similar to the LSTM, $i$ is the input gate, $f$ is the forget gate, and $o$ is the output gate. $\bar{c}_t^m$ is the proposed update to the hybrid memory ${c}^m_t \in \mathbb{R}^{d^{m}_{mem}}$ at time $t$. ${h}^m_t \in \mathbb{R}^{d^{m}_{mem}}$ is the time distributed output of each modality. 

The neural \inter \ dynamics code $z_{t}$ is the output of the \mab \ at the previous time-step and is discussed in detail in next subsection. This neural \inter \ dynamics code $z_{t}$ is passed to each of the individual \tnames s\ and is the hybrid factor, allowing each individual \tnames \ to carry \inter \ dynamics that it finds related to its modality. The set of weights $W^m_*$,$U^m_*$ and $V^m_*$ respectively map the input of \tnames \ $x^m_t$, output of \tnames \ $h^m_t$, and neural \inter \ dynamics code $z_{t}$ to each \tnames \ memory space using affine transformations.

\makeatletter
\def\BState{\State\hskip-\ALG@thistlm}
\makeatother

\begin{algorithm}[b!]
\caption{\pipeline \ (\pipelines), \ourl \ (\ours) and \mab \ (\mabs) Formulation}\label{alg_LSTHM}
\begin{algorithmic}[1]
\Procedure{$\textrm{\pipelines}$}{$\mathbf{X}^m$}
\State $c_0,h_0,z_0 \gets \mathbf{0}$
\For {$t = 1, ..., T$}:
\State $h_t \gets \textrm{LSTHM\_Step} (\bigcup_{m \in M} \{ x^m_t\}, z_{t-1})$
\State $z_t \gets \textrm{\mabs\_Step} (h_t)$
\EndFor
\Return $h_T, z_T$
\EndProcedure

\Procedure{$\textrm{LSTHM\_Step}$}{$\bigcup_{m \in M} \{{x}^m_t\}$, $z_{t-1}$}
\For {$m \in M$}: \ \ $\triangleleft$ for all the $M$ modalities
\State $i_t^m \gets \sigma(W_i^m\ x^m_t+U^m_i\ h^m_{t-1}+V^m_i\ z_{t-1}+b^m_{i})$ 
\State $f^m_t \gets \sigma(W^m_{f}\ x^m_t + U^m_{f}\ h^m_{t-1} + V^m_f\ z_{t-1}+b^m_{f})$ 
\State $o^k_t \gets \sigma(W^m_{o}\ x^m_t + U^m_{o}\ h^m_{t-1} + V^m_o\ z_{t-1}+b^m_{o})$ 
\State $\bar{c}_t^m \gets W_{\bar{c}}^m\ x^m_t + U_{\bar{c}}^m\ h^m_{t-1} + V_{\bar{c}}^m\ z_{t-1} + b^m_{\bar{c}}$
\State ${c}^m_t \gets f^m_t \odot {c}^m_{t-1} + i^m_t \odot tanh(\bar{c}_t^m)$
\State $h^m_t \gets o^m_t \odot tanh({c}^m_t)$
\EndFor
\State $h_t \gets \bigoplus_{m \in M} h^m_t $ \\
\noindent \quad \ \ \Return $h_t$
\EndProcedure

\Procedure{$\textrm{\mabs\_Step}$}{$h_t$}
\State $a_t \gets \mathcal{A}(h_t; \theta_{\mathcal{A}})$ $\triangleleft$ $K$ output coefficients
\State $\widetilde{h}_t \gets a_t \odot \langle \Uparrow_K h_t \rangle$ 
\For {$m \in M$}: $\triangleleft$ calculate \inter \ dynamics
\State $s^m_{t} \gets \mathcal{C}_m (\widetilde{h}^m_{t}; \theta_{\mathcal{C}_m})$ 
\EndFor
\State $s_t \gets \bigoplus_{m \in M} s^m_{t}$ 
\State $z_t \gets \mathcal{G}({s_t; \theta_{\mathcal{G}}})$

\noindent \quad \ \ \Return $z_t$
\EndProcedure
\end{algorithmic}
\end{algorithm}

\subsection{\mab}
\label{subsec:mab}
At each timestamp $t$, various \inter \ dynamics across the modalities can occur simultaneously. For example, the first instance can be the connection between a smile and positive phrase both happening at time $t$. A second instance can be the occurrence of the same smile at time $t$ being connected to an excited voice at time $t-4$, that was carried to time $t$ using the audio \tnames \ memory. In both of these examples, \inter \ dynamics exist at time $t$. Therefore, not only do \inter \ dynamics span across various modalities, they are scattered across time forming asynchronous \inter \ dynamics.

The \mab \ is a network that can capture multiple different, possibly asynchronous, \inter \ dynamics and encode all of them in a neural \inter \ dynamics code $z_t$. In the most important step of the \mab, different dimensions of \tnames \ outputs $h^m_t$ are assigned attention coefficients according to whether or not they form \inter \ dynamics. These attention coefficients will be high if the dimension contributes to formation of a \inter \ dynamics and low if they are irrelevant. The coefficient assignment is performed multiple times due to the existence of possibly multiple such \inter \ dynamics across the outputs of \tnames. The \mab \ is formulated in Algorithm 1. We assume a maximum of $K$ \inter \ dynamics to be present at each timestamp $t$. To obtain the $K$ attention coefficients, $K$ softmax distributions are assigned to the concatenated \tnames \ memories using a deep neural network $\mathcal{A} : \mathbb{R}^{{d_{mem}}} \mapsto \mathbb{R}^{K \times {d_{mem}}}$ with ${d_{mem}} = \sum_{m \in M} {d^{m}_{mem}}$. At each timestep $t$, the output of LSTHM is the set $\{h^m_t : m \in M, h^m_t \in \mathbb{R}^{{d^m_{mem}}}\}$. $\mathcal{A}$ takes the concatenation of \tnames \ outputs $h_t = \bigoplus_{m \in M} h^m_t, h_t \in \mathbb{R}^{{d_{mem}}}$ as input and outputs a set of $K$ attentions $\{a^k_{t} : k \leq K, \ a^k_{t} \in \mathbb{R}^{{d_{mem}}}\}$ with $a_t = \bigoplus_{k=1}^K a^k_t$, $a_t \in \mathbb{R}^{K \times {d_{mem}}}$. $\mathcal{A}$ has a softmax layer at top of the network which takes the softmax activation along each one of the $K$ dimensions of its output $a_t$. As a result, $a^k_{t} \ge 0, \sum_{i=1}^{{d_{mem}}} (a^k_{t})_i  = 1$ which forms a probability distribution over the output dimensions. $h_t$ is then broadcasted (from $\mathbb{R}^{{d_{mem}}}$ to $\mathbb{R}^{K \times {d_{mem}}}$) and element-wise multiplied by the $a_t$ to produce attended outputs $\widetilde{h}_t=\{\widetilde{h}^k_{t} : k \leq K \ , \widetilde{h}^k_{t} \in \mathbb{R}^{{d_{mem}}} \}$, $\widetilde{h}_t \in \mathbb{R}^{K \times {d_{mem}}}$. $\Uparrow_K$ denotes broadcasting by parameter $K$.

\begin{figure}[t!]
\centering{
\includegraphics[width=0.9\linewidth]{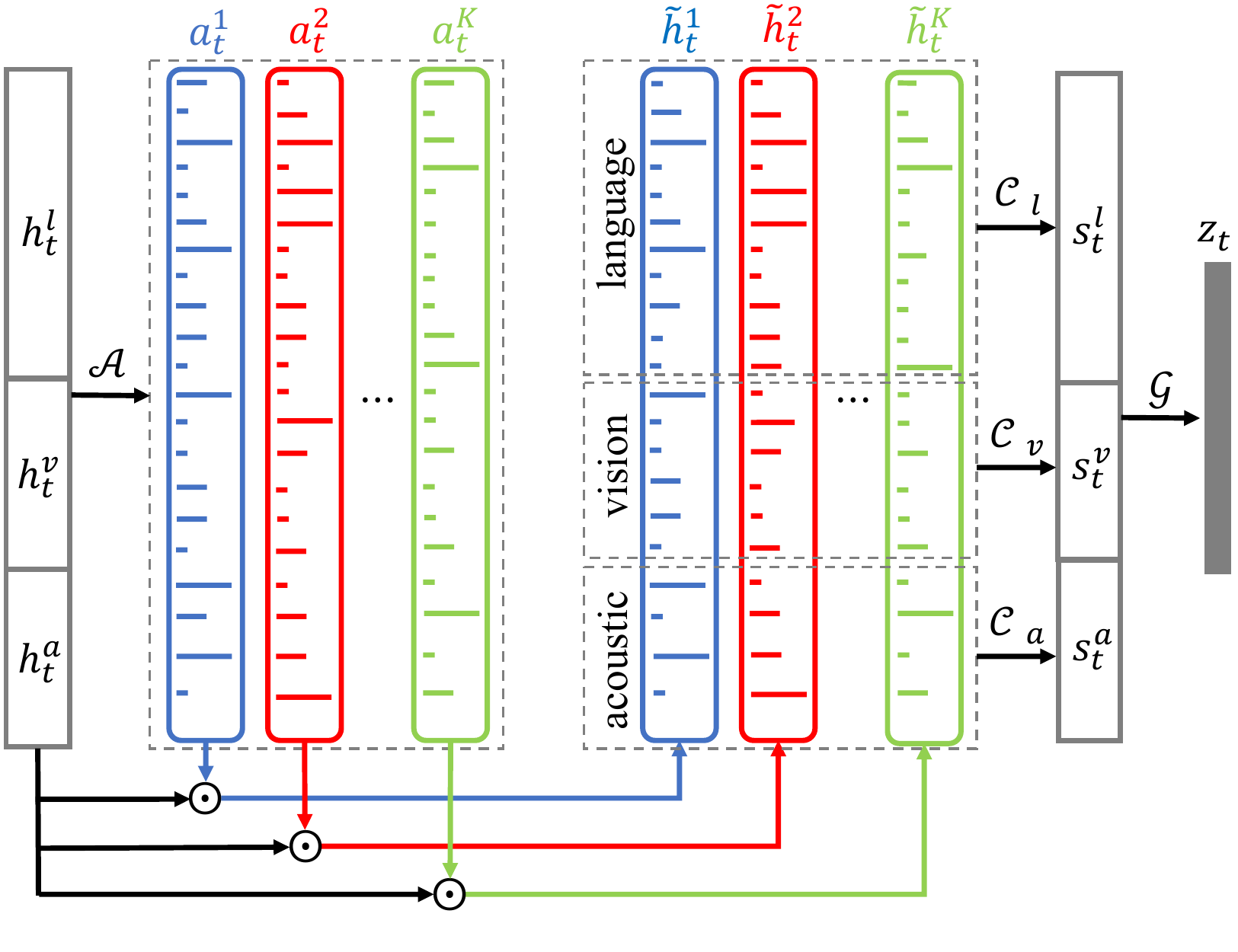}}
    \caption{Overview of \mab \ (\mabs).}
    \label{fig:overview}
\end{figure}

The first dimension of $\widetilde{h}_t$ contains information needed for the first \inter \ dynamic highlighted using $a^1_t$, the second dimension of $\widetilde{h}_t$ contains information for the second \inter \ dynamic using $a^2_t$, and so on until $K$. $\widetilde{h}_t$ is high dimensional but ideally considered sparse due to presence of dimensions with zero value after element-wise multiplication with attentions. Therefore, $\widetilde{h}_t$ is split into $m$ different parts -- one for each modality $m$ -- and undergoes dimensionality reduction using $\mathcal{C}_m : \mathbb{R}^{K \times {d^m_{mem}}} \mapsto \mathbb{R}^{{d^m_{local}}}, \forall m \in M$ with ${d^m_{local}}$ as the target low dimension of each modality split in $\widetilde{h}_t$. The set of networks $\{ \mathcal{C}_m : m \in M \}$ maps the attended outputs of each modality $\widetilde{h}^m_t$ to the same vector space. This dimensionality reduction produces a dense code $s^m_t$ for the $K$ times attended dimensions of each modality. Finally, the set of all $M$ attended modality outputs, $s_t = \bigoplus_{m \in M} s^m_t$, are passed into a deep neural network $\mathcal{G} : \mathbb{R}^{\sum_{m \in M} {{d^m_{local}}}} \mapsto \mathbb{R}^{{d_{mem}}}$ to generate the neural \inter \ dynamics code $z_t$ at time $t$.

\section{Experimental Methodology}
In this paper we benchmark \pipelines 's\ understanding of human communication on three tasks: 1) multimodal sentiment analysis, 2) multimodal speaker traits recognition and 3) multimodal emotion recognition. We perform experimentations on six publicly available datasets and compare the performance of \pipelines \ with the performance of state-of-the-art approaches on the same datasets. To ensure generalization of the model, all the datasets are split into train, validation and test sets that include no identical speakers between sets, i.e. all the speakers in the test set are different from the train and validation sets. All models are re-trained on the same train/validation/test splits. To train the \pipelines \ for different tasks, the final outputs $h_T$ and neural \inter \ dynamics code $z_T$ are the inputs to another deep neural network that performs classification (categorical cross-entropy loss function) or regression (mean squared error loss function). The code, hyperparameters and instruction on data splits are publicly available at \url{https://github.com/A2Zadeh/MARN}. 

Following is the description of different benchmarks.  
\subsection{Multimodal Sentiment Analysis}
\quad \textbf{CMU-MOSI} The CMU-MOSI dataset \cite{zadeh2016multimodal} is a collection of 2199 opinion video clips. Each opinion video is annotated with sentiment in the range [-3,3]. There are 1284 segments in the train set, 229 in the validation set and 686 in the test set.

\textbf{ICT-MMMO} The ICT-MMMO dataset \cite{wollmer2013youtube} consists of online social review videos that encompass a strong diversity in how people express opinions, annotated at the video level for sentiment. The dataset contains 340 multimodal review videos, of which 220 are used for training, 40 for validation and 80 for testing.

\textbf{YouTube} The YouTube dataset \cite{morency2011towards} contains videos from the social media web site YouTube that span a wide range of product reviews and opinion videos. Out of 46 videos, 30 are used for training, 5 for validation and 11 for testing.

\textbf{MOUD} To show that \pipelines \ is generalizable to other languages, we perform experimentation on the MOUD dataset \cite{perez-rosas_utterance-level_2013} which consists of product review videos in Spanish. Each video consists of multiple segments labeled to display positive, negative or neutral sentiment. Out of 79 videos in the dataset, 49 are used for training, 10 for validation and 20 for testing.

\subsection{Multimodal Speaker Trait Recognition}
\quad  \textbf{POM} Persuasion Opinion Multimodal (POM) dataset \cite{Park:2014:CAP:2663204.2663260} contains movie review videos annotated for the following speaker traits: confidence, passion, dominance, credibility, entertaining, reserved, trusting, relaxed, nervous, humorous and persuasive. 903 videos were split into 600 were for training, 100 for validation and 203 for testing. 

\subsection{Multimodal Emotion Recognition}
\quad  \textbf{IEMOCAP} The IEMOCAP dataset \cite{Busso2008IEMOCAP:Interactiveemotionaldyadic} consists of 151 videos of recorded dialogues, with 2 speakers per session for a total of 302 videos across the dataset. Each segment is annotated for the presence of 9 emotions (angry, excited, fear, sad, surprised, frustrated, happy, disappointed and neutral) as well as valence, arousal and dominance. The dataset is recorded across 5 sessions with 5 pairs of speakers. To ensure speaker independent learning, the dataset is split at the level of sessions: training is performed on 3 sessions (6 distinct speakers) while validation and testing are each performed on 1 session (2 distinct speakers).

\subsection{Multimodal Computational Descriptors}
All the datasets consist of videos where only one speaker is in front of the camera. The descriptors we used for each of the modalities are as follows: 

\textbf{Language} All the datasets provide manual transcriptions. We use pre-trained word embeddings (glove.840B.300d) \cite{pennington2014glove} to convert the transcripts of videos into a sequence of word vectors. The dimension of the word vectors is 300. 

\textbf{Vision} Facet \cite{emotient} is used to extract a set of features including per-frame basic and advanced emotions and facial action units as indicators of facial muscle movement.

\textbf{Acoustic} We use COVAREP \cite{degottex2014covarep} to extract low level acoustic features including 12 Mel-frequency cepstral coefficients (MFCCs), pitch tracking and voiced/unvoiced segmenting features, glottal source parameters, peak slope parameters and maxima dispersion quotients.

\textbf{Modality Alignment} To reach the same time alignment between different modalities we choose the granularity of the input to be at the level of words. The words are aligned with audio using P2FA \cite{P2FA} to get their exact utterance times. Time step $t$ represents the $t$th spoken word in the transcript. We treat speech pause as a word with vector values of all zero across dimensions. The visual and acoustic modalities follow the same granularity. We use expected feature values across the entire word for vision and acoustic since they are extracted at a higher frequency (30 Hz for vision and 100 Hz for acoustic). 

\subsection{Comparison Metrics}
Different datasets in our experiments have different labels. For binary classification and multiclass classification we report accuracy A$^C$ where $C$ denotes the number of classes, and F1 score. For regression we report Mean Absolute Error MAE and Pearson's correlation $r$. For all the metrics, higher values denote better performance, except MAE where lower values denote better performance. 

\subsection{Baseline Models}
\label{sec:base}
We compare the performance of our \pipelines \ to the following state-of-the-art models in multimodal sentiment analysis, speaker trait recognition, and emotion recognition. All baselines are trained for datasets for complete comparison. 

\textbf{TFN} (Tensor Fusion Network) \cite{tensoremnlp17} explicitly models \intra \ and \inter \ dynamics by creating a multi-dimensional tensor that captures unimodal, bimodal and trimodal interactions across three modalities. It is the current state of the art for CMU-MOSI dataset.

\textbf{BC-LSTM} (Bidirectional Contextual LSTM) \cite{contextmultimodalacl2017} is a model for context-dependent sentiment analysis and emotion recognition, currently state of the art on the IEMOCAP and MOUD datasets.

\textbf{MV-LSTM} (Multi-View LSTM) \cite{rajagopalan2016extending} is a recurrent model that designates special regions inside one LSTM to different views of the data. 

\textbf{C-MKL} (Convolutional Neural Network (CNN) with Multiple Kernel Learning) \cite{DBLP:conf/emnlp/PoriaCG15} is a model which uses a CNN for visual feature extraction and multiple kernel learning for prediction.

\textbf{THMM} (Tri-modal Hidden Markov Model) \cite{morency2011towards} performs early fusion of the modalities by concatenation and uses a HMM for classification.

\textbf{SVM} (Support Vector Machine) \cite{Cortes:1995:SN:218919.218929} a SVM is trained on the concatenated multimodal features for classification or regression \cite{zadeh2016multimodal,perez-rosas_utterance-level_2013,Park:2014:CAP:2663204.2663260}. To compare to another strong non-neural baseline we use \textbf{RF} (Random Forest) \cite{Breiman:2001:RF:570181.570182} using similar multimodal inputs.

\textbf{SAL-CNN} (Selective Additive Learning CNN) \cite{wang2016select} is a model that attempts to prevent identity-dependent information from being learned by using Gaussian corruption introduced to the neuron outputs. 

\textbf{EF-HCRF}: (Hidden Conditional Random Field) \cite{Quattoni:2007:HCR:1313053.1313265} uses a HCRF to learn a set of latent variables conditioned on the concatenated input at each time step. We also implement the following variations: 1) \textbf{EF-LDHCRF} (Latent Discriminative HCRFs) \cite{morency2007latent} are a class of models that learn hidden states in a CRF using a latent code between observed concatenated input and hidden output. 2) \textbf{MV-HCRF}: Multi-view HCRF \cite{song2012multi} is an extension of the HCRF for Multi-view data, explicitly capturing view-shared and view specific sub-structures. 3) \textbf{MV-LDHCRF}: is a variation of the MV-HCRF model that uses LDHCRF instead of HCRF. 4) \textbf{EF-HSSHCRF}: (Hierarchical Sequence Summarization HCRF) \cite{song2013action} is a layered model that uses HCRFs with latent variables to learn hidden spatio-temporal dynamics. 5) \textbf{MV-HSSHCRF}: further extends \textbf{EF-HSSHCRF} by performing Multi-view hierarchical sequence summary representation. The best performing early fusion model is reported as \textbf{EF-HCRF$_{(\star)}$} while the best multi-view model is reported as \textbf{MV-HCRF$_{(\star)}$}, where $\star \in \{ \textrm{h, l, s}\}$ to represent HCRF, LDCRF and HSSCRF respectively.

\textbf{DF} (Deep Fusion) \cite{Nojavanasghari:2016:DMF:2993148.2993176} is a model that trains one deep model for each modality and performs decision voting on the output of each modality network. 

\begin{table}[t!]
\fontsize{7}{10}\selectfont
\centering
\setlength\tabcolsep{5.3pt}
\begin{tabular}{l c c c c c}
\Xhline{3\arrayrulewidth}
 & \multicolumn{2}{c}{{Binary}} & \multicolumn{1}{c}{{Multiclass}} & \multicolumn{2}{c}{{Regression}} \\
\cmidrule(lr{.5em}){2-3} \cmidrule(lr{.5em}){4-4} \cmidrule(lr{.5em}){5-6}
Method       & A$^{2}$ & F1 & A$^{7}$ & MAE & Corr\\ \Xhline{0.5\arrayrulewidth}
Majority       & 50.2 &   50.1   &	 17.5	&     1.864 &  0.057  \\
RF         & 56.4 &  56.3    &  21.3 	&		-     	&  - \\ 
SVM-MD     & 71.6 &   72.3   & 26.5  	& 1.100     &  0.559 \\ 
THMM		& 50.7	& 45.4	& 17.8& - & -\\
SAL-CNN    & 73.0 &   -      &  -	&  -        	& -\\ 
C-MKL          & 72.3 &   72.0   &  30.2	&  -       		& -\\
EF-HCRF$_{(\star)}$		& 65.3$_{\textrm{(h)}}$ & 65.4$_{\textrm{(h)}}$ & 24.6$_{\textrm{(l)}}$ & - & -\\
MV-HCRF$_{(\star)}$	& 65.6$_{\textrm{(s)}}$ & 65.7$_{\textrm{(s)}}$ & 24.6$_{\textrm{(l)}}$ & - & -\\
DF              & 72.3 &   72.1   &  26.8 	& 1.143     &  0.518 \\
EF-LSTM$_{(\star)}$		& 73.3$_{\textrm{(sb)}}$ &   73.2$_{\textrm{(sb)}}$   & 32.4$_{\textrm{(-)}}$ &  1.023$_{\textrm{(-)}}$  & 0.622$_{\textrm{(-)}}$ \\
MV-LSTM			& 73.9 &   74.0   &   33.2	&  1.019    &  0.601 \\
BC-LSTM         & 73.9 &   73.9   &  28.7	& 1.079    	&  0.581 \\ 
TFN             & 74.6 &   74.5   &  28.7	&    1.040   	& 0.587   \\ \Xhline{0.5\arrayrulewidth}
{\pipelines } (no MAB)       	& {76.5} & {76.5} &  30.8	&  {0.998}   	& 0.582 \\ 
{\pipelines } (no $\mathcal{A}$)  & 59.3$_{(3)}$ & 36.0$_{(3)}$ & 22.0$_{(3)}$ & 1.438$_{(5)}$ & 0.060$_{(5)}$\\
{\pipelines }      		& \textbf{77.1$_{(4)}$}	& \textbf{77.0$_{(4)}$} & \textbf{34.7$_{(3)}$} & \textbf{0.968$_{(4)}$}  & \textbf{0.625$_{(5)}$} \\  \Xhline{0.5\arrayrulewidth}
Human                   & 85.7 &   87.5   &  53.9	&    0.710    	&  0.820  \\ \Xhline{3\arrayrulewidth}
\end{tabular}
\caption{Sentiment prediction results on CMU-MOSI test set using multimodal methods. Our model outperforms the previous baselines and the best scores are highlighted in bold.}
\label{table:mosi}
\end{table}

\begin{table}[tb]
\fontsize{7}{10}\selectfont
\centering
\setlength\tabcolsep{3pt}
\begin{tabular}{l c c c c c c}
\Xhline{3\arrayrulewidth}
& \multicolumn{2}{c}{{ICT-MMMO Binary}} & \multicolumn{2}{c}{{YouTube Multiclass}} & \multicolumn{2}{c}{{MOUD Binary}} \\
 \cmidrule(lr{.5em}){2-3}  \cmidrule(lr{.5em}){4-5} \cmidrule(lr{.5em}){6-7} 
Method        	& A$^{2}$ & F1 & A$^{3}$ & F1 & A$^{2}$ & F1 \\ \Xhline{0.5\arrayrulewidth}
Majority		& 40.0 & 22.9 & 42.4 & 25.2 & 60.4 & 45.5\\
RF				& 70.0 & 69.8 & 49.3 & 49.2 & 64.2 & 63.3\\
SVM     		& 68.8 & 68.7 & 42.4 & 37.9 & 60.4 & 45.5\\
THMM			& 53.8	& 53.0 & 42.4 & 27.9 & 58.5 & 52.7\\
C-MKL			& 80.0 & 72.4 & 50.2 & 50.8 & 74.0 & 74.7 \\
EF-HCRF$_{(\star)}$	& 81.3$_{\textrm{(l)}}$ & 79.6$_{\textrm{(l)}}$ & 45.8$_{\textrm{(l)}}$ & 45.0$_{\textrm{(l)}}$ & 54.7$_{\textrm{(h)}}$ & 54.7$_{\textrm{(h)}}$ \\
MV-HCRF$_{(\star)}$	& 81.3$_{\textrm{(l)}}$ & 79.6$_{\textrm{(l)}}$ & 44.1$_{\textrm{(s)}}$ & 44.0$_{\textrm{(s)}}$	& 60.4$_{\textrm{(l)}}$ & 47.8$_{\textrm{(l)}}$ \\
DF   			&  77.5	&	77.5 & 45.8 & 32.0 & 67.0 & 67.1 \\
EF-LSTM$_{(\star)}$	& 80.0$_{\textrm{(sb)}}$ & 78.5$_{\textrm{(sb)}}$ & 44.1$_{\textrm{(-)}}$ & 43.6$_{\textrm{(-)}}$ & 67.0$_{\textrm{(-)}}$ & 64.3$_{\textrm{(-)}}$ \\
MV-LSTM			& 72.5 & 72.3  & 45.8 & 43.3 & 57.6 & 48.2 \\
BC-LSTM    		& 70.0	&	70.1 & 47.5 & 47.3 & 72.6 & 72.9 \\ 
TFN      		& 72.5	&	72.6 & 47.5	& 41.0 & 63.2 & 61.7 \\ \Xhline{0.5\arrayrulewidth}
{\pipelines } (no MAB)	& 82.5 & 82.4 & 47.5 & 42.8 & 75.5 & 72.9 \\ 
{\pipelines } (no $\mathcal{A}$)  & 80.0$_{(5)}$ & 79.1$_{(5)}$ & 44.1$_{(5)}$ & 29.3$_{(5)}$ & 63.2$_{(5)}$ & 61.9$_{(5)}$ \\
{\pipelines }	&  \textbf{86.3$_{(2)}$} & \textbf{85.9$_{(2)}$} & \textbf{54.2$_{(6)}$} & \textbf{52.9$_{(6)}$} & \textbf{81.1$_{(2)}$} & \textbf{81.2$_{(2)}$} \\ \Xhline{3\arrayrulewidth}
\end{tabular}
\caption{Sentiment prediction results on ICT-MMMO, YouTube and MOUD test sets. Our model outperforms the previous baselines and the best scores are highlighted in bold.}
\label{table:mmmo+you+moud}
\end{table}

\begin{table*}[tb]
\fontsize{7}{10}\selectfont
\centering
\setlength\tabcolsep{6.3pt}
\begin{tabular}{l c c c c c c c c c c c c c c c c c c c c c c c c}
\Xhline{3\arrayrulewidth}
Task & \multicolumn{1}{c}{Confident} & \multicolumn{1}{c}{Passionate} & \multicolumn{1}{c}{Dominant} & \multicolumn{1}{c}{Credible} & \multicolumn{1}{c}{Entertaining} & \multicolumn{1}{c}{Reserved} & \multicolumn{1}{c}{Trusting} & \multicolumn{1}{c}{Relaxed} & \multicolumn{1}{c}{Nervous} & \multicolumn{1}{c}{Persuasive} & \multicolumn{1}{c}{Humorous}\\
Method & A$^{7}$ & A$^{7}$ & A$^{7}$ & A$^{7}$ & A$^{7}$ & A$^{5}$ & A$^{5}$ & A$^{5}$ & A$^{5}$ & A$^{7}$ & A$^{5}$ \\ \Xhline{0.5\arrayrulewidth}
Majority		& 19.2 & 20.2 & 18.2 & 21.7  & 19.7 & 29.6 & 44.3 & 39.4 &  24.1 & 20.7 & 6.9 \\
SVM     		&  26.6 & 20.7 & 35.0 & 25.1 & 31.5 & 34.0 & 50.2 & 49.8 & 41.4 & 28.1 & 36.0 \\
RF     			&  26.6 & 27.1 & 26.1 & 23.2 & 26.1 & 34.0 & 53.2 & 40.9 & 36.0 & 25.6 & 40.4  \\
THMM    		&  24.1&15.3&29.1&27.6&12.3&22.7&31.0&31.5&27.1&17.2&24.6 \\
DF   			& 25.6&24.1&34.0&26.1&29.6&30.0&53.7&50.2&42.4&26.6&34.5\\
EF-LSTM$_{(\star)}$	& 25.1$_{\textrm{(b)}}$&30.5$_{\textrm{(sb)}}$&36.9$_{\textrm{(s)}}$&29.6$_{\textrm{(b)}}$&33.5$_{\textrm{(b)}}$&33.5$_{\textrm{(sb)}}$&52.7$_{\textrm{(sb)}}$&48.3$_{\textrm{(-)}}$&44.8$_{\textrm{(sb)}}$&25.6$_{\textrm{(sb)}}$&39.4$_{\textrm{(b)}}$ \\
MV-LSTM 		& 25.6 & 28.6 & 34.5 & 25.6 & 29.1 & 33.0 & 52.2 & 50.7 & 42.4 &  26.1 & 38.9\\
BC-LSTM    		&   26.6&26.6&33.0&27.6&29.6&33.0&52.2&47.3&36.0&27.1&36.5\\ 
TFN      		& 	24.1&31.0&34.5&24.6&29.1&30.5&38.9&35.5&42.4&27.6&33.0\\  \Xhline{0.5\arrayrulewidth}
{\pipelines } (no MAB) 	& 26.1&27.1&35.5&28.1&30.0&32.0&55.2&50.7&42.4&29.1&33.5\\ 
{\pipelines } (no $\mathcal{A}$)  & 24.6$_{(6)}$ & 32.0$_{(5)}$ & 34.0$_{(5)}$ & 24.6$_{(6)}$ & 29.6$_{(6)}$ & 32.5$_{(6)}$ & 53.2$_{(6)}$ & 49.3$_{(6)}$ & 42.4$_{(5)}$ & 29.6$_{(6)}$ & 42.4$_{(4)}$ \\
{\pipelines } 	&  
\textbf{29.1$_{(2)}$}&\textbf{33.0$_{(6)}$}&\textbf{38.4$_{(6)}$}&\textbf{31.5$_{(2)}$}&\textbf{33.5$_{(3)}$}&\textbf{36.9$_{(1)}$}&\textbf{55.7$_{(1)}$}&\textbf{52.2$_{(6)}$}&\textbf{47.3$_{(5)}$}&\textbf{31.0$_{(3)}$}&\textbf{44.8$_{(5)}$} \\ 
\Xhline{3\arrayrulewidth}
\end{tabular}
\caption{Speaker personality trait recognition results on POM test set. Our model outperforms the previous baselines and the best scores are highlighted in bold.}
\label{table:pom-att}
\end{table*}

\textbf{EF-LSTM} (Early Fusion LSTM) concatenates the inputs from different modalities at each time-step and uses that as the input to a single LSTM. We also implement the Stacked, (\textbf{EF-SLSTM}) Bidirectional (\textbf{EF-BLSTM}) and Stacked Bidirectional (\textbf{EF-SBLSTM}) LSTMs for stronger baselines. The best performing model is reported as \textbf{EF-LSTM$_{(\star)}$}, $\star \in \{ \textrm{-, s, b, sb}\}$ denoting vanilla, stacked, bidirectional and stacked bidirectional LSTMs respectively.

\textbf{Majority} performs majority voting for classification tasks, and predicts the expected label for regression tasks. This baseline is useful as a lower bound of model performance. 

\textbf{Human} performance is calculated for CMU-MOSI dataset which offers per annotator results. This is the accuracy of human performance in a one-vs-rest classification/regression. 

Finally, \textbf{\pipelines}\ indicates our proposed model. Additionally, the modified baseline \textbf{\pipelines \ (no \mabs)} removes the \mabs \ and learns no dense \inter \ dynamics code $z$. This model can be seen as three disjoint LSTMs and is used to investigate the importance of modeling temporal \inter \ dynamics. The next modified baseline \textbf{\pipelines \ (no $\mathcal{A}$)} removes the $\mathcal{A}$ deep network and sets all $K$ attention coefficients $a^k_t = 1$ ($h^k_t = \tilde{h}^k_t$). This comparison shows whether explicitly outlining the \inter \ dynamics using the attention coefficients is required. For \pipelines \ and \pipelines \ (no $\mathcal{A}$), $K$ is treated as a hyperparamter and the best value of $K$ is indicated in parenthesis next to the best reported result.

\section{Experimental Results}

\subsection{Results on CMU-MOSI dataset}
We summarize the results on the CMU-MOSI dataset in Table \ref{table:mosi}. We are able to achieve new state-of-the-art results for this dataset in all the metrics using the \pipelines. This highlights our model's capability in understanding sentiment aspect of multimodal communication.

\subsection{Results on ICT-MMMO, YouTube, MOUD Datasets}
We achieve state-of-the-art performance with significant improvement over all the comparison metrics for two English sentiment analysis datasets. Table \ref{table:mmmo+you+moud} shows the comparison of our \pipelines \ with state-of-the-art approaches for ICT-MMMO dataset as well as the comparison for YouTube dataset. To assess the generalization of the \pipelines \ to speakers communicating in different languages, we compare with state-of-the-art approaches for sentiment analysis on MOUD, with opinion utterance video clips in Spanish. The final third of Table \ref{table:mmmo+you+moud} shows these results where we also achieve significant improvement over state-of-the-art approaches.

\subsection{Results on POM Dataset}
We experiment on speaker traits recognition based on observed multimodal communicative behaviors. Table \ref{table:pom-att} shows the performance of the \pipelines \ on POM dataset, where it achieves state-of-the-art accuracies on all 11 speaker trait recognition tasks including persuasiveness and credibility. 

\subsection{Results on IEMOCAP Dataset} 
Our results for multimodal emotion recognition on IEMOCAP dataset are reported in Table \ref{table:iemocap}. Our approach achieves state-of-the-art performance in emotion recognition: both emotion classification as well as continuous emotion regression except for the case of correlation in dominance which our results are competitive but not state of the art. 

\begin{table}[tb]
\fontsize{7}{10}\selectfont
\centering
\setlength\tabcolsep{0.16pt}
\begin{tabular}{l c c c c c c c c}
\Xhline{3\arrayrulewidth}
Task       		& \multicolumn{2}{c}{Emotions} & \multicolumn{2}{c}{Valence} & \multicolumn{2}{c}{Arousal} & \multicolumn{2}{c}{Dominance} \\
Method        	& A$^{9}$ & F1 & MAE & Corr & MAE & Corr & MAE & Corr  \\ \Xhline{0.5\arrayrulewidth}
Majority		& 21.2 & 7.4	& 2.042 & -0.02 & 1.352 & 0.01 & 1.331 & 0.17	\\
SVM     		& 24.1 & 18.0 & 0.251 & 0.06 & 0.546 & 0.54 & 0.687 & 0.42 \\
RF     			& 27.3 & 25.3 &- & - &- &- &- &- \\
THMM			& 23.5 & 10.8 & - & - &- &- &- &- \\
C-MKL			& 34.0 & 31.1 & - & - & - & - & - & -  \\
EF-HCRF$_{(\star)}$			& 32.0$_{\textrm{(s)}}$ & 20.5$_{\textrm{(s)}}$ & - & - &- &- &- &- \\
MV-HCRF$_{(\star)}$			& 32.0$_{\textrm{(s)}}$ & 20.5$_{\textrm{(s)}}$ & - & - &- &- &- &- \\
DF   			& 26.1 & 20.0 & 0.250 & -0.04 & 0.613 & 0.27 & 0.726 & 0.09 \\
EF-LSTM$_{(\star)}$	& 34.1$_{\textrm{(s)}}$ & 32.3$_{\textrm{(s)}}$ & 0.244$_{\textrm{(-)}}$ & 0.09$_{\textrm{(-)}}$ & 0.512$_{\textrm{(b)}}$ & 0.62$_{\textrm{(-)}}$ & 0.669$_{\textrm{(s)}}$ & 0.51$_{\textrm{(sb)}}$\\
MV-LSTM   		& 31.3 & 26.7 & 0.257 & 0.02 & 0.513 & 0.62 & 0.668 & \textbf{0.52}\\
BC-LSTM    		& 35.9 & 34.1 & 0.248 & 0.07 & 0.593 & 0.40 & 0.733 & 0.32 \\ 
TFN      		& 36.0	& 34.5 & 0.251 & 0.04 & 0.521 & 0.55 & 0.671 & 0.43 \\  \Xhline{0.5\arrayrulewidth}
{\pipelines } (no MAB)		&  31.2 & 28.0 & 0.246 & 0.09 & 0.509 & 0.63 & 0.679 & 0.44\\ 
{\pipelines } (no ${\mathcal{A}}$) 	& 23.0$_{(3)}$ & 10.9$_{(3)}$ & 0.249$_{(5)}$ & 0.05$_{(5)}$ & 0.609$_{(4)}$ & 0.29$_{(4)}$ & 0.752$_{(4)}$ & 0.21$_{(5)}$ \\
{\pipelines }			& \textbf{37.0}$_{(4)}$ & \textbf{35.9}$_{(4)}$ & \textbf{0.242}$_{(6)}$ & \textbf{0.10}$_{(5)}$ & \textbf{0.497}$_{(3)}$ & \textbf{0.65}$_{(3)}$ & \textbf{0.655}$_{(1)}$ & 0.50$_{(5)}$\\ \Xhline{3\arrayrulewidth}
\end{tabular}
\caption{Emotion recognition results on IEMOCAP test set using multimodal methods. Our model outperforms the previous baselines and the best scores are highlighted in bold.}
\label{table:iemocap}
\end{table}


\section{Discussion}
Our experiments indicate outstanding performance of \pipelines \ in modeling various attributes related to human communication. In this section, we aim to better understand different characteristics of our model.

\subsection{Properties of Attentions}
To better understand the effects of attentions, we pose four fundamental research questions (RQ) in this section as RQ1: \pipelines \ (no MAB): whether the \inter \ dynamics are helpful. RQ2: \pipelines \ (no $\mathcal{A}$): whether the attention coefficients are needed. RQ3: \pipelines: whether one attention is enough to extract all \inter \ dynamics. RQ4: whether different tasks and datasets require different numbers of attentions.

\textbf{RQ1:} \pipelines \ (no \mabs) model can only learn simple rules among modalities such as decision voting or simple co-occurrence rules such as Tensor Fusion baseline. Across all datasets, \pipelines \ (no MAB) is outperformed by \pipelines. This indicates that continuous modeling of \inter \ dynamics is crucial in understanding human communication. 

\textbf{RQ2:} Whether or not the presence of the coefficients $a_t$ are crucial is an important research question. From the results tables, we notice that the \pipelines \ (no $\mathcal{A}$) baseline severely under-performs compared to \pipelines. This supports the importance of the attentions in the \mabs. Without these attentions, \pipelines \ is not able to accurately model the \inter \ dynamics.

\textbf{RQ3:} In our experiments the \pipelines \ with only one attention (like conventional attention models) under-performs compared to the models with multiple attentions. One could argue that the models with more attentions have more parameters, and as a result their better performance may not be due to better modeling of \inter \ dynamics, but rather due to more parameters. However we performed extensive grid search on the number of parameters in \pipelines \ with one attention. Increasing the number of parameters further (by increasing dense layers, \tnames \ cellsizes etc.) did not improve performance. This indicates that the better performance of \pipelines \ with multiple attentions is not due to the higher number of parameters but rather due to better modeling of \inter \ dynamics.

\textbf{RQ4:} Different tasks and datasets require different number of attentions. This is highly dependent on each dataset's nature and the underlying interconnections between modalities. 

\subsection{Visualization of Attentions}

\begin{figure}[t!]
\centering{
\includegraphics[width=0.92\linewidth]{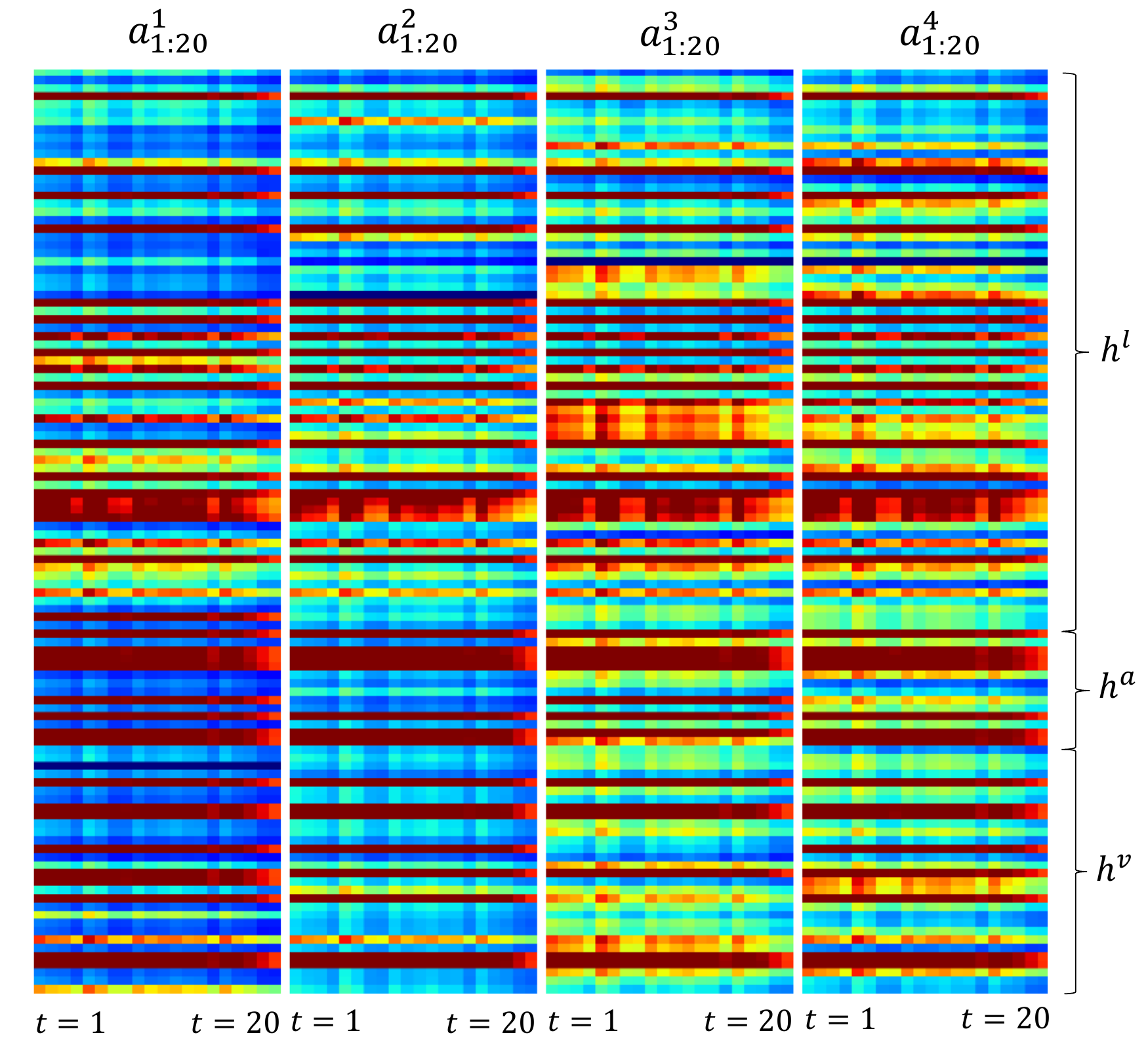}}
\caption{Visualization of attention units throughout time. Blue: activated attentions and red: non-activated attentions. The learned attentions are diverse and evolve across time.}
\label{fig:attv}
\end{figure}

We visually display how each attention is sensitive to different dimensions of \tnames \ outputs in Figure \ref{fig:attv}. Each column of the figure denoted by $a^k$ shows the behavior of the $k$th attention on a sample video from CMU-MOSI. The left side of $a^k$ is $t=1$ and the right side is $t=20$, since the sequence has 20 words. The $y$ axis shows what modality the dimension belongs to. Dark blue means high coefficients and red means low coefficients. Our observations (O) are detailed below:

\textbf{O1:} By comparing each of the attentions together, they show diversity on which dimensions they are sensitive to, indicating that each attention is sensitive to different \inter \ dynamics. 

\textbf{O2:} Some attention coefficients are not active (always red) throughout time. These dimensions carry only \intra \ dynamics needed by that modality and not other modalities. Hence, they are not needed for \inter \ dynamics and will carry no weight in their formation. 

\textbf{O3:} Attentions change their behaviors across time. For some coefficients, these changes are more drastic than the others. We suspect that the less drastic the change in an attention dimension over time, the higher the chances of that dimension being part of multiple \inter \ dynamics. Thus more attentions activate this important dimension.

\textbf{O4:} Some attentions focus on \inter \ dynamics that involve only two modalities. For example, in $a^3$, the audio modality has no dark blue dimensions, while in $a^1$ all the modalities have dark blue dimensions. The attentions seem to have residual effects. $a^1$ shows activations over a broad set of variables while $a^4$ shows activation for fewer sets, indicating that attentions could learn to act in a complementary way.

\section{Conclusion}
In this paper we modeled multimodal human communication using a novel neural approach called the \pipeline \ (\pipelines). Our approach is designed to model both \intra \ dynamics as well as \inter \ dynamics continuously through time. View-specific dynamics are modeled using a \tname \ (\tnames) for each modality. Various \inter \ dynamics are identified at each time-step using the \mab \ (\mabs) which outputs a multimodal neural code for the hybrid memory of \tnames. \pipelines \ achieves state-of-the-art results in 6 publicly available datasets and across 16 different attributes related to understanding human communication.

\section{Acknowledgements}
This project was partially supported by Oculus research grant. We thank the reviewers for their valuable feedback.


%

\bibliographystyle{aaai}
{\small\bibliography{citations.bib}}

\end{document}